\documentclass{article}

\usepackage{microtype}
\usepackage{graphicx}
\usepackage{subfigure}
\usepackage{booktabs} 

\usepackage{hyperref}

\usepackage[accepted]{icml2019}

\icmltitlerunning{Achieving Fairness in Determining Medicaid Eligibility through Fairgroup Construction}

\begin{document}

\twocolumn[
\icmltitle{Achieving Fairness in Determining Medicaid Eligibility through Fairgroup Construction}



\icmlsetsymbol{equal}{*}

\begin{icmlauthorlist}
\icmlauthor{Boli Fang}{iu}
\icmlauthor{Miao Jiang}{iu}
\icmlauthor{Jerry Shen}{iu}
\end{icmlauthorlist}

\icmlaffiliation{iu}{Indiana University, Bloomington, Indiana, USA}

\icmlcorrespondingauthor{Boli Fang}{bfang@iu.edu}

\icmlkeywords{Machine Learning, ICML}

\vskip 0.3in
]



\printAffiliationsAndNotice{}

\begin{abstract}
Effective complements to human judgment, artificial intelligence techniques have started to aid human decisions in complicated social problems across the world. In the context of United States for instance, automated ML/DL classification models offer complements to human decisions in determining Medicaid eligibility. However, given the limitations in ML/DL model design, these algorithms may fail to leverage various factors for decision making, resulting in improper decisions that allocate resources to individuals who may not be in the most need. In view of such an issue, we propose in this paper the method of \textit{fairgroup construction}, based on the legal doctrine of \textit{disparate impact}, to improve the fairness of regressive classifiers. Experiments on American Community Survey dataset demonstrate that our method could be easily adapted to a variety of regressive classification models to boost their fairness in deciding Medicaid Eligibility, while maintaining high levels of classification accuracy. 
\end{abstract}

\section{Introduction}
\label{intro}

As defined by the United Nations Sustainable Development Goals, social decision problems in equality, fairness, and sustainability are top priorities for developed and developing nations across the world. In particular, proper allocation of health and medical resources are vital for the well-being of citizens across different countries. While the majority of endeavors in previous work centered on the developing world, one cannot ignore the related issues in developed countries. According to the American Community Survey \cite{us_census_bureau}, millions of American households are regularly receiving governmental assistance in receiving Medicaid, a compensation scheme designated for low-income individuals to receive proper reimbursement for necessary medical treatment. It is noted in the same dataset that over 16 million households in America are living "below poverty level", yet a substantial amount of poor households are not yet receiving Medicaid. On the other hand, out of the households that are receiving Medicaid, a highly non-trivial amount - around 56$\%$ - of these households do not live under poverty. Such great disparity behooves the researchers to introduce a complementary decision maker that better takes various factors of the problem into consideration, and recent advancements in Machine Learning and Deep Learning algorithms have offered objective insights into these problems \cite{morse_2018}.   


However, given the limitations of ML/DL algorithms, the issue of fairness has also been the focus for a lot of current machine learning research. Taking into consideration aspects of computational actions and socioeconomic context, previous researchers have focused on two subcategories of fairness as benchmarks - outcome fairness and process fairness. Given the nature of most social welfare programs, which are designed to maximize the interests of individuals and households with low socioeconomic status, outcome fairness is often more important than process fairness.

Moreover, some factors are more important than others when discussing fairness. In the context of Medicaid eligibility, for instance, it is important to include as many individuals living under poverty into the program as possible, while minimizing the number of individuals that do not need such assistance so as to allow for the optimal allocation of the finite monetary and health resources. 

Thus, given such considerations, we introduce in this paper a novel method for regressive classification algorithm to more fairly distribute Medicaid resources among individuals. Given an agnostic classifier which might produce biased classification results, we construct fairgroups in the testing data set, and proceed to classify the entire testing set by first classifying representatives of fairgroups and then propagating the decision to other data points. Here, the notion of fairness follows that of \textit{disparate impact} \cite{feldman2015certifying}, which calls for similar levels of representation for all the groups of people in different decision outcome classes. Our contributions in this work can be summarized as follows: 

\begin{enumerate}
\item We introduce a method to help regressive classifiers to better allocate Medicaid resources by constructing \textbf{fairgroups}, and achieves outcome fairness in the Medicaid Decision Problem with respect to the features that we hope to impose fairness on. 
\item Our algorithm also takes into consideration other features not involved in defining fairness while making decisions on fairness, so that individuals with similar features will be classified in similar ways.
\item The method to achieve fairness in our paper is easily adaptable to other decision making procedures, such as judicial verdicts, acceptance to educational programs and approval of credit card applications.  
\end{enumerate}

\section{Related Work}
\label{related}

Previous work on fairness in machine learning can be largely divided into two groups. The first group has centered on the mathematical definition and existence of fairness \cite{feldman2015certifying, Zafar, chierichetti2017fair}. Along this track, alternative measures such as statistical parity, disparate impact, and individual fairness \cite{chierichetti2017fair} have been produced. Additionally, Grgic-Hlaca et. al. (\citeyear{grgic2016case}) covers common notions of fairness and introduces methods of measuring fairness such as feature-apriori fairness, feature-accuracy fairness, and feature-disparity fairness. \cite{kleinberg2016inherent} suggested that although it's not possible to achieve some desired properties of fairness at the same time, including "protected" features in algorithms would increase the equity and efficiency of machine learning models. 

The second group has centered on algorithms to achieve fairness. Along the route of disparate impact, \cite{feldman2015certifying} has described algorithms to spot the presence of disparate impact through Support Vector Machine, while \cite{chierichetti2017fair} applied the notion of disparate impact to design an algorithm that achieves balance in unsupervised clustering algorithms. \cite{chierichetti2017fair} also introduces the notion of \textit{protected and unprotected features}, which we have used in our paper.

\section{Model}
In this section we present a novel strategy called \textit{fair-grouping} to achieve fairness in classification results. This strategy adopts the notion of fairness as related to \textit{disparate impact} \cite{feldman2015certifying}, where practices based on neutral rules and laws may still more adversely affect individuals with one protected feature than those without.

\subsection{Preliminaries}
We first define the terminology to be used in subsequent description. A \textit{protected feature} is a feature that carries special importance and is of priority when making relevant decisions. An \textit{unprotected feature}, on the other hand, is of relative minor importance in decision making. Since the problem in our paper primarily focuses on discrete label classification with discrete features, we assume, without loss of generality and for sake of simplicity, that the protected traits are binary and that the classification label class is also binary. Given a protected feature $A$ along with the dataset, the \textit{balance} $B$ of the dataset with respect to $A$ is defined as 
$$
Bal(A) = \min\{\frac{\#\{A=0\}}{\#\{A=1\}},\frac{\#\{A=1\}}{\#\{A=0\}}\} \in [0,1],
$$
where $Bal(A) = 0$ refers to the case of all data points having the same feature value of $A$, and $Bal(A) = 1$ refers to the case where $\#\{A=0\}=\#\{A=1\}$.
A dataset is $\alpha$-\textit{fair} with respect to feature $A$ if the balance of $A$ does not go below a certain number $\alpha\in[0,1]$. In other words, a dataset is $\alpha$-disparate with respect to $A$ if the groups with 2 different values in $A$ have a bounded and relative balanced numerical ratio between $\frac{1}{\alpha}$ and $\alpha$. Following the doctrine of disparate impact as stated in \cite{feldman2015certifying}, we say that a classification is $(\alpha,i)$-fair if the group corresponding to label $i$ in the classification class $L = \{+,-\}$ is $\alpha$-fair, meaning that the protected feature is fairly represented with balance at least $\alpha$ in group $i$. 

\subsection{Fair-group construction}
We provide in this section the details of the algorithms we will use to achieve fairness in classification. Assume that we already have a classifier $C$ which yields predictions for data points and might not yield $\alpha$-fair classification results. Overall, our algorithm constructs fair-groups from testing data, and conducts classification on the data points with $C$ while taking the properties of the fairgroups into consideration.

The sections below provide more details of our method. 

\subsubsection{Correlation Computation}

Most of the social decision problems involve different features of varying degrees of relevance and importance to the goal. To achieve this goal, we compute the correlation coefficient between feature $X_i$ and the outcome $Y$ to determine the contribution of each feature to the final classification outcome:
\[Corr(X_i,Y) = \frac{E[X_iY] - E[X_i]E[Y]}{\sqrt{Var(X_i)Var(Y)}}.\]

We then rank all the features by an increasing order of the absolute values of correlation coefficients, because higher correlation values indicate greater statistical significance in either positive or negative directions. Then, we assign to each feature $X_i$ a weight $w_i$ which is equal to the rank by increasing values of the correlation coefficients. The weight $w_i$ reflects the significance of feature $X_i$ in the classifier. 

After constructing the relative weight $w_i$ of each feature $X_i$ from the correlation coefficients, we examine the actual values of $X_i$ for each data point $j$, here denoted by $x_{ij}$. If a feature $X_i$ is positively correlated with $Y$, then we rank all data by the decreasing order of the corresponding $x_{ij}$'s of the feature $X_i$, and define $r_{ij}$ as the rank of $x_{ij}$ in the set of all values of $X_i$'s. Alternatively, if a feature has negative correlation, the the data is ranked in increasing order of $x_{ij}$, and $r_{ij}$'s are defined accordingly. Intuitively, the rank $r_{ij}$'s show how much influence each feature $X_i$ in data point $j$ has to the final classification prediction. These ranks are constructed in a way to make sure that the data points with higher values of $X_i$ are given enough consideration, since higher feature values in socialogical datasets are often likely to correspond to special cases requiring extra attention.  

Finally, for each attribute $X_i$ in corresponding to data point $j$, we define $r'_{ij} = w_ir_{ij}$ as the \textit{feature importance index}, and define $\mathbf{r}'_j$ as the \textit{feature importance vector} corresponding to data point $j$. The feature importance vector reveals information about the relative importance of data point $j$, and such information will be used to construct fairgroups for subsequent fair classification.


\subsubsection{Fairgroup construction}
With each data point now represented in the form of feature importance vectors, we now examine how close these data points are in terms of the influence each data point might exert to the final classification outcome, and how data points with similar features can be grouped together for easier analysis. To achieve these goals, we define a suitable distance between two vectors and consider a clustering problem where similar data points are grouped together.     

Notice that each of the entries in the feature importance vectors are integers corresponding to different rankings, and that closer ranks imply similarity in one feature. Thus, we make use of the Manhattan-L1 distance to describe the distance between feature importance vectors $\mathbf{r}'_p, \mathbf{r}'_q$:
$$
d(\mathbf{r}'_p, \mathbf{r}'_q) = \sum_{i=1}^{N} |r'_{ip} - r'_{iq}| = \sum_{i=1}^N w_i|r_{ip} - r_{iq}|,
$$
Here $N$ refers to the number of unprotected features.

Afterwards, we consider a $k$-median cluster algorithm to divide the entire dataset into $k$ groups, each containing points with similar feature values. Within each cluster, we look at the protected features. Without loss of generality, we assume that the protected feature is binary, and that our goal is to maintain the balance of the protected feature $A$ does not go below a certain threshold $t$. Since this requirement implies that the ratio between $\#\{A=0\}$ and $\#\{A=1\}$ falls between $t$ and $\frac{1}{t}$, we match as many $A=0$ and $A=1$ data points as possible on condition that the ratio between $\#\{A=0\}$ and $\#\{A=1\}$ in each match falls between $t$ and $1/t$. A set consisting of data points in such matches is denoted as a \textit{fairgroup}.    

\subsubsection{Classification with respect to each fairgroup}
For each fair-group we have thus constructed, we randomly pick a point to be classified by $C$. If the point is labeled as $+$, we apply the same label to all other data points in the group. Alternatively, if the point is labeled as $-$, we need to take into consideration the properties of the protected feature to determine whether other data points in the same fair-group will be given the same label. For instance, in the case of Food Stamp distribution, protected features such as poverty should be treated as a protected feature only in the positive label class, because our primary goal is to ensure that people receiving food stamps are mainly composed of people living under the poverty threshold. On the other hand, for decision problems that favor similar representation of one feature in different label classes, we need to include the feature in both positive and negative classes. While determining admission eligibility for admission into selective schools, for instance, it is important that the odds of being admitted and rejected are roughly the same across different demographic groups to ensure equality.

Moreover, to reduce the negative effect of potential misclassification as much as possible, we construct as many fairgroups as possible by first expressing $t$ and $\frac{1}{t}$ as ratios $\frac{p}{q}$ and $\frac{q}{p}$, where $p,q$ are co-prime integers. Starting from $\frac{\#\{A=0\}}{\#\{A=1\}}$, we iteratively match $p$ data points where $A=0$ with $q$ data points where $A=1$(or $q$ data points where $A=0$ with $p$ data points where $A=1$) depending on whether $\frac{p}{q}$ or $\frac{q}{p}$ is smaller than and closer to the ratio of unmatched $\frac{\#\{A=0\}}{\#\{A=1\}}$. These matched $p+q$ points will form a fairgroup, and corresponding numbers of $A=0$, $A=1$ points will be moved from the unmatched point set. We repeat the procedure until all the points are matched or unmatchable.This procedure ensures that we create maximal numbers of fairgroups, so that even when one fairgroup is misclassified due to the misclassification of the randomly drawn point, the effects on the overall fairness and consistency can be minimal. 

\section{Experiments}

\subsection{Dataset}
\label{dataset}
To conduct experiments using the model explained above, we use the United States Census American Community Survey data. Consisting over 2 million entries, the individual level microdata displays important features, including status of receiving Medicaid for a specific household. 

\subsubsection{Protected Features}

The feature importance scores have been calculated using the correlation formula in section 3.2 with respect to the training data. Other variables include disability, number of persons in a household, poverty status, locations, etc. The numerical values of these features are listed in table 1. For this experiment, we have selected \textbf{household income} and \textbf{poverty status} as protected variables because they have the highest importance of the model. To make \textbf{household income} an indicator variable, we have set an experimental threshold of \$20000, and define those households earning below the threshold as \textit{households to be protected}. 

\begin{table}[h]
\label{sample-table}
\vskip 0.1in
\begin{center}
\begin{small}
\begin{sc}
\begin{tabular}{lc}
\toprule
Feature & Feature importance\\
\midrule
Age    &  0.0783\\
Division    &  0.00532\\
Region         & 0.00132\\
State         &   0.00197\\
Gender            &  0.00215\\
Number of Children & 0.00306\\
Hearing Difficulty & 0.0121\\
Vision Difficulty & 0.0121\\
Ambulatory difficulty & 0.0121\\\textbf{}
Self-care difficulty & 0.0121\\
Class of workers & 0.127\\
\textbf{Household Income }& \textbf{0.398}\\
Interest Income & 0.111\\
Race & 0.00587 \\
\textbf{Poverty Status} & \textbf{0.1747}\\
\bottomrule
\end{tabular}
\end{sc}
\end{small}
\end{center}
\caption{Feature importance of Medicaid Dataset}
\vskip -0.1in
\end{table}

\subsubsection{Target Variable}
Here in our experiments, the target variable is the feature which indicates whether a single individual has finally received medicaid or not. This is a binary feature with two options 'yes' and 'no'. 

\subsection{Results}
We have carried out two sets of experiments to show that our algorithm is able to improve the fairness in the predictive results, as compared to pure regressive classifiers such as logistic regression. By the description of our method, we cluster all household data points into 5 clusters by K-median clustering\cite{zhu2015brain}. In each cluster, we maintain the same ratio for poverty and non-poverty households by setting the balance as $\frac{8}{2} = \frac{4}{1}$ between poverty and non-poverty households, so as to impose a 80\% poverty percentage among the people receiving MedicAid.

Table 2 and 3 list the experimental results for different regressive classifiers when the protected features are \textbf{household income} and \textbf{poverty status} respectively. We have experimented on Linear Regression, Logistic Regression and Support Vector Machine, three of the most representative regression models, to demonstrate the effectiveness of our method. We notice that for all three models, our fairgroup construction effectively boosts the level of protected features in fairness, increasing the proportion of poverty by 15 to 20 \%. At the same time, the classification accuracy of the respective models remains very high and comparable to the original models. This indicates that the clustering step in our algorithm preserves the similarity between data points in classification.


\begin{table}[h]
\label{sample-table}
\vskip 0.15in
\begin{center}
\begin{small}
\begin{sc}
\begin{tabular}{lcc}
\toprule
Method & \% of Poverty &  Accuracy\\
\midrule
Logistic Regression & 67.4 & 92.6\\ 
Linear Regression & 65.3 & 90.2\\
SVM & 68.7 & 91.5 \\
Logistic + Fairgroup& 84.3 & 89.5 \\
Linear Rgression + Fairgroup & 82.7 & 88.1\\
SVM + Fairgroup & 83.1 & 88.3\\ 

\bottomrule
\end{tabular}
\end{sc}
\end{small}
\end{center}
\caption{Experiment results on Medicaid with Household Income as Protected Feature}
\vskip -0.1in
\end{table}

\begin{table}[h]
\label{sample-table}
\vskip 0.15in
\begin{center}
\begin{small}
\begin{sc}    
\begin{tabular}{lcc}
\toprule
Method & \% of Poverty & Accuracy\\
\midrule
Logistic Regression &   67.4 & 92.6\\ 
Linear Regression & 65.3 & 90.2\\
SVM & 68.7 & 91.5 \\
Logistic + Fairgroup & 84.7 & 89.3 \\
Linear Rgression + Fairgroup & 83.4 & 86.9\\
SVM + Fairgroup & 83.6 & 88.9\\ 
\bottomrule
\end{tabular}
\end{sc}
\end{small}
\end{center}
\caption{Experimental results on Medicaid with Poverty Level as Protected Feature}
\vskip -0.1in
\end{table}

\section{Conclusion}
In this work we present a novel approach to solve the problem of Medicaid Eligibility Determination through classifiers that achieve fairness in outcome. To achieve our goal, we propose the strategy of \textit{fair-group} construction, to promote representation of households in poverty in the group of people receiving Medicaid. Experiments on the US Census individual level microdata yields results that are more consistent among samples with similar attributes. As a part of our future work. we hope to apply our method to address the current social problems related to inequality and inequity in both the developed and developing world.

\bibliography{main}

\begin{thebibliography}{8}
\providecommand{\natexlab}[1]{#1}
\providecommand{\url}[1]{\texttt{#1}}
\expandafter\ifx\csname urlstyle\endcsname\relax
  \providecommand{\doi}[1]{doi: #1}\else
  \providecommand{\doi}{doi: \begingroup \urlstyle{rm}\Url}\fi

\bibitem[Bureau()]{us_census_bureau}
Bureau, U.~C.
\newblock American community survey 2017 5-year estimate.
\newblock URL \url{https://www.census.gov/programs-surveys/acs/?}

\bibitem[Chierichetti et~al.(2017)Chierichetti, Kumar, Lattanzi, and
  Vassilvitskii]{chierichetti2017fair}
Chierichetti, F., Kumar, R., Lattanzi, S., and Vassilvitskii, S.
\newblock Fair clustering through fairlets.
\newblock In \emph{Advances in Neural Information Processing Systems}, pp.\
  5029--5037, 2017.

\bibitem[Feldman et~al.(2015)Feldman, Friedler, Moeller, Scheidegger, and
  Venkatasubramanian]{feldman2015certifying}
Feldman, M., Friedler, S.~A., Moeller, J., Scheidegger, C., and
  Venkatasubramanian, S.
\newblock Certifying and removing disparate impact.
\newblock In \emph{Proceedings of the 21th ACM SIGKDD International Conference
  on Knowledge Discovery and Data Mining}, pp.\  259--268. ACM, 2015.

\bibitem[Grgic-Hlaca et~al.(2016)Grgic-Hlaca, Zafar, Gummadi, and
  Weller]{grgic2016case}
Grgic-Hlaca, N., Zafar, M.~B., Gummadi, K.~P., and Weller, A.
\newblock The case for process fairness in learning: Feature selection for fair
  decision making.
\newblock In \emph{NIPS Symposium on Machine Learning and the Law}, volume~1,
  pp.\ ~2, 2016.

\bibitem[Kleinberg et~al.(2016)Kleinberg, Mullainathan, and
  Raghavan]{kleinberg2016inherent}
Kleinberg, J., Mullainathan, S., and Raghavan, M.
\newblock Inherent trade-offs in the fair determination of risk scores.
\newblock \emph{arXiv preprint arXiv:1609.05807}, 2016.

\bibitem[Morse(2018)]{morse_2018}
Morse, S.
\newblock Artificial intelligence helps insurers identify medicare members who
  also qualify for medicaid, Nov 2018.

\bibitem[Zafar et~al.(2017)Zafar, Valera, Rogriguez, and Gummadi]{Zafar}
Zafar, M.~B., Valera, I., Rogriguez, M.~G., and Gummadi, K.~P.
\newblock {Fairness Constraints: Mechanisms for Fair Classification}.
\newblock In \emph{Proceedings of the 20th International Conference on
  Artificial Intelligence and Statistics}, Proceedings of Machine Learning
  Research, pp.\  962--970. PMLR, 2017.

\bibitem[Zhu \& Shi(2015)Zhu and Shi]{zhu2015brain}
Zhu, H. and Shi, Y.
\newblock Brain storm optimization algorithms with k-medians clustering
  algorithms.
\newblock In \emph{2015 Seventh International Conference on Advanced
  Computational Intelligence (ICACI)}, pp.\  107--110. IEEE, 2015.

\end{thebibliography}
\bibliographystyle{icml2019}

\end{document}